\pdfoutput=1

\documentclass[11pt]{article}

\usepackage[final]{acl}
\usepackage{tabularx} 
\usepackage{amsmath}
\usepackage{upgreek}
\usepackage{times}
\usepackage{latexsym}
\usepackage{multirow}

\usepackage[T1]{fontenc}

\usepackage[utf8]{inputenc}

\usepackage{microtype}

\usepackage{inconsolata}

\usepackage{graphicx}

\title{Is My Text in Your AI Model? \\Gradient-based Membership Inference Test applied to LLMs}

\author{
    Gonzalo Mancera\textsuperscript{1}, 
    Daniel DeAlcala\textsuperscript{1}, 
    Julian Fierrez\textsuperscript{1}, 
    Ruben Tolosana\textsuperscript{1}, 
    Aythami Morales\textsuperscript{1} \\
    \textsuperscript{1}Biometrics and Data Pattern Analytics Lab, Universidad Autónoma de Madrid, Spain \\
    \texttt{\{gonzalo.mancera, daniel.dealcala, julian.fierrez,} \\ 
    \texttt{ruben.tolosana, aythami.morales\}@uam.es}
}

\begin{document}
\maketitle
\begin{abstract}
This work adapts and studies the gradient-based Membership Inference Test (gMINT) to the classification of text based on LLMs. MINT is a general approach intended to determine if given data was used for training machine learning models, and this work focuses on its application to the domain of Natural Language Processing. Using gradient-based analysis, the MINT model identifies whether particular data samples were included during the language model training phase, addressing growing concerns about data privacy in machine learning. The method was evaluated in seven Transformer-based models and six datasets comprising over 2.5 million sentences, focusing on text classification tasks. Experimental results demonstrate MINT's robustness, achieving AUC scores between 85\% and 99\%, depending on data size and model architecture. These findings highlight MINT's potential as a scalable and reliable tool for auditing machine learning models, ensuring transparency, safeguarding sensitive data, and fostering ethical compliance in the deployment of AI/NLP technologies.
\end{abstract}

\section{Introduction}
In recent years, Natural Language Processing (NLP) technologies have been increasingly deployed in various sectors such as healthcare \cite{locke2021natural}, legal \cite{katz2023natural}, customer service \cite{olujimi2023nlp}, education \cite{shaik2022review}, or finance \cite{du2025natural}, profoundly changing the way information is analyzed and understood. Advancements in machine learning have enabled the processing of vast textual data, driving rapid progress in NLP. However, these developments have also raised important ethical and legal concerns that require immediate attention. In response, the European Union introduced new legislation in June 2024\footnote{\url{https://artificialintelligenceact.eu}}, specifically aimed at regulating the use of AI technologies. The proposed legislation represents a unified effort to address the ethical and social implications associated with the widespread use of AI in general and NLP systems in particular. Its primary goal is to protect the fundamental rights of citizens through measures that ensure transparency and accountability in the development and implementation of AI technologies. That legal framework and related international guidelines enforce the creation of audit tools to oversee the responsible integration of AI/NLP systems into society, while also gaining a deeper understanding of their functioning and associated risks \cite{USA}. 

In this scenario of raising privacy concerns in AI deployments \cite{privacy2025}, we focus our attention on Membership Inference Attacks (MIAs), which involve adversaries attempting to extract sensitive information about the data used to train a model \cite{shokri2017membership}. Research has demonstrated the feasibility of these attacks and has proposed various strategies to address them. Building on this foundation, \citet{dealcala2024my} introduced an innovative framework called the Membership Inference Test (MINT), designed to detect unauthorized data use when training AI/ML models, which was soon followed by a gradient-based variant, gMINT \cite{gMINT2025}. The primary objective of MINT approaches (including gMINT) is to determine whether specific data have been used without the necessary permissions. (A MINT demonstrator was recently released by \citet{dealcala2025mintdemo}.\footnote{\url{https://ai-mintest.org/}})
This approach seeks to audit AI and machine learning models, improving transparency and explainability \cite{arrieta2020explainable, liao2023ai}. It enables individuals to identify whether certain data, such as personal information or other sensitive data \cite{sensitivenets2021}, were used as training data. Although MIAs and MINT share similar objectives, they operate in different contexts and address different aspects of the problem.

\begin{figure}[t!]

\centering
\includegraphics[width=\columnwidth]{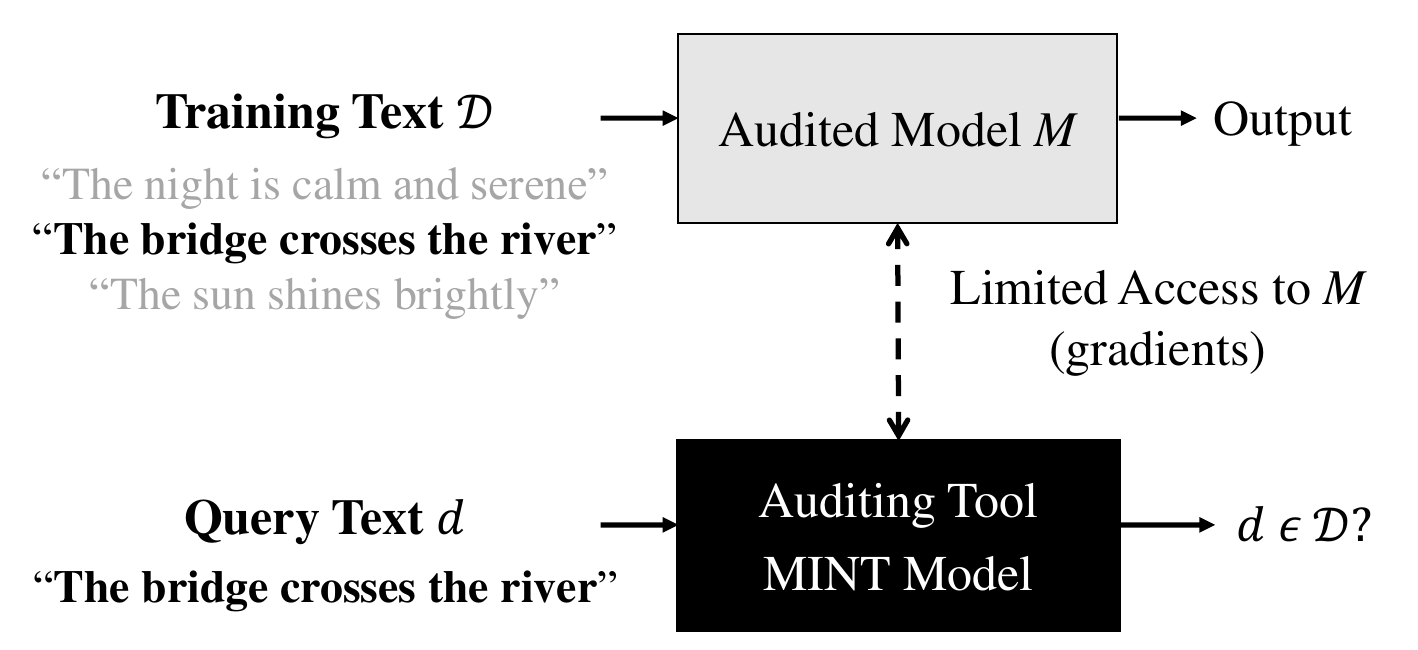}
\caption{The objective of MINT is to discern whether given data (\textit{d}) were utilized in the training process of an AI Model (\textit{M}) trained with a specific database (\(\mathcal{D}\)).}
\label{fig:esquemamint}
\end{figure}

This paper addresses the challenge of identifying whether a given text was utilized in the training of a Language Model (see Fig. \ref{fig:esquemamint}). The study seeks to further emphasize the versatility and applicability of the general MINT approach, initially validated for images by \citet{dealcala2024my}, by demonstrating its effectiveness in language-related tasks. The main contributions are as follows.
\begin{itemize}
    \item We adjust and apply MINT for the first time in the literature to LLMs, in particular the gradient-based variant (gMINT) proposed by \citet{gMINT2025}. Using the gradients generated during the learning process, the gMINT model takes advantage of the patterns related to how each training sample influences the model's training and optimization.
    \item We evaluated gMINT in the context of text classification. The assessment includes experiments with seven Transformer-based Language Models and six datasets comprising 2.5 million sentences spanning diverse topics.
    \item The findings demonstrate that gMINT effectively identifies which samples belong to the training process with high accuracy, showcasing its ability to accurately trace the data used for model development.
\end{itemize}


\section{Related Works}

\subsection{Membership Inference Attacks (MIAs) in Language Models}
Membership Inference Attacks (MIAs) aim to determine whether specific data samples were included in the training set of a machine learning model. These attacks exploit subtle behavioral differences in how a model processes training data compared to unseen data, such as variations in loss trajectories \cite{liu2022membership}, similarity between distributions \cite{gao2023similarity}, or activation patterns \cite{hu2022membership}. However, executing MIAs is inherently challenging because most of the samples encountered by a model have not been used during training. This creates an imbalance between the data that the model has seen and the large amount of unseen data, making it difficult for an attacker to reliably infer membership \cite{watson2021importance}.

For Language Models, MIAs face unique challenges due to the large volume and diversity of text data used during training, as well as the overlapping distributions between training and unseen data \cite{yao2024survey}. Existing methods, such as those proposed by \cite{mattern2023membership}, leverage loss comparisons with similar sentences generated via masked language models. Other approaches, such as user inference techniques, compare the likelihood of user-specific samples to determine membership \cite{kandpal2023user}. Although effective, these methods are computationally expensive and often require extensive auxiliary data or shadow models, limiting their scalability. Furthermore, the scale and single-epoch nature of the language model pre-training reduce the visibility of training data, complicating membership inference \cite{duan2024membership}. The complexity of the language also creates challenges for attackers trying to identify discernible patterns. These factors make MIAs in language models especially complex, requiring advanced techniques to detect vulnerabilities while preserving the effectiveness and security of the model \cite{mireshghallah2022quantifying}.

\subsection{Membership Inference Test (MINT)}

MINT is a Membership Inference Test designed to assess the compliance of machine learning models with current regulatory frameworks governing the use of training data. The process requires a certain level of cooperation from the model owner to ensure that the model's development and operation are aligned with legal and ethical guidelines. The central goal of MINT is to create robust methodologies that are applicable to various data modalities, such as text, images, and audio, while accounting for different levels of transparency and information shared by the model developer regarding the model's training and operational procedures \cite{dealcala2024my}. These scenarios range from situations where little or no information is available to cases where the details of the model are fully disclosed \cite{dealcala2024comprehensive}.

Within the MINT framework, auditors are tasked with evaluating the model, having access either to the original model or to limited details about its structure and training processes. This aligns with the practical application of MINT, particularly in responding to evolving legal requirements and advancing data protection measures. Importantly, MINT distinguishes itself from traditional Membership Inference Attacks (MIAs), which typically involve creating shadow models to mimic the behavior of the original model. In contrast, MINT enables auditors to apply inference techniques directly to the original model, leading to different techniques and more accurate detection results. Several factors, such as the level of access to the model or the type of data modality, can influence the effectiveness and results of MINT, making it adaptable to a variety of compliance challenges \cite{dealcala2024comprehensive}.

To the best of our knowledge, MINT has not been studied in the context of language models. MINT can be an important auditing tool to enhance the transparency and trustworthiness of current and future language models.

\begin{figure*}[h!]
\centering
\includegraphics[width=\textwidth]{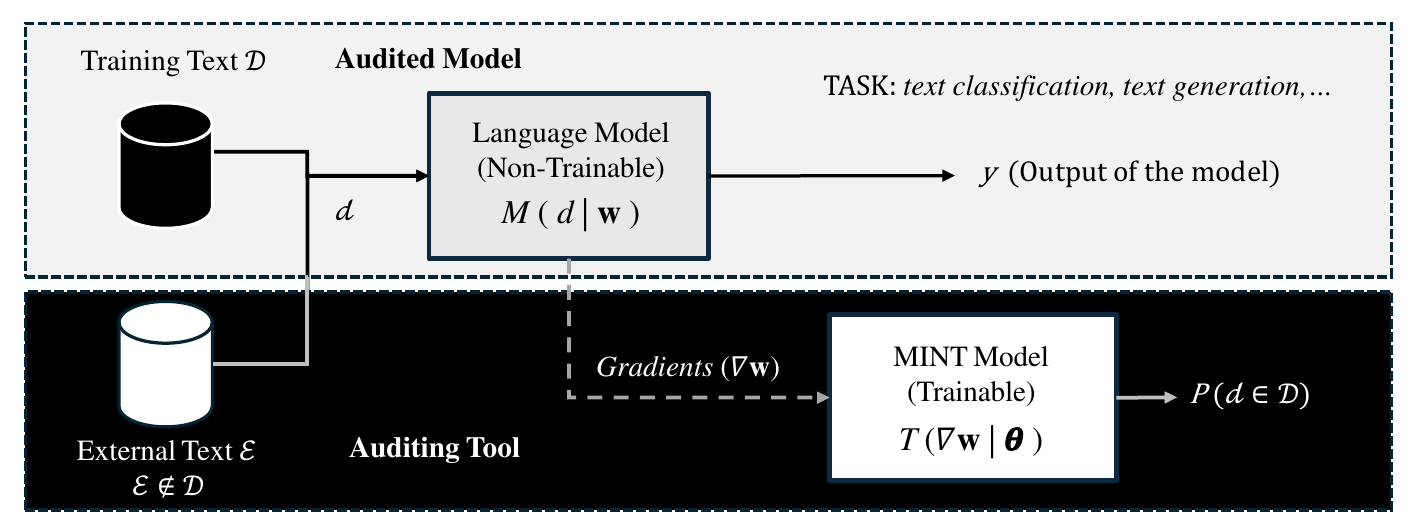}
\caption{The MINT Model $T(\nabla \textbf{w} | \boldsymbol\uptheta)$ is trained to determine whether a given data instance $d$ was part of the training data $\mathcal{D}$ used to develop a Language Model \textit{M}. The inputs for the MINT Model consist of gradients $\nabla \textbf{w}$, specifically the gradients of the model’s parameters with respect to the data samples $d$.}
\label{fig:esquemamint_ext}
\end{figure*}

\section{Membership Inference Test: Gradient-based Approach (gMINT)}

The original MINT proposal \cite{dealcala2024my} was tested on Face Recognition models (FR). Here we study and develop MINT for Language Models, specifically models trained for text classification. Language models exhibit several substantial differences with respect to image models: while image models typically return an embedding in a representation space, language models output a sequence of tokens or a probability distribution over the vocabulary. Consequently, their training methods and the type of information they output are markedly dissimilar.

Gradients can play a crucial role in MINT. When a data sample belongs to the training set, the model may react to it in a specific way, particularly reflected in the gradients, which in general will have distinct characteristics compared to gradients associated with data outside the training set. This behavior was leveraged by \citet{gMINT2025} generating the so-called gradient-based MINT (gMINT).

\subsection{Problem Statement}\label{problem}

Given a training dataset \(\mathcal{D}\) and an external dataset \(\mathcal{E}\), we define a set of samples \( d \) that belong to \( \mathcal{D} \cup \mathcal{E} \). We assume the existence of a model \( M \) that has been trained on \( \mathcal{D} \) for a specific task. For each input sample \( d \), the model \( M \) produces an output \( y \) as a function of \( d \) and a set of parameters \( \textbf{w} \) learned during the training process. This dependency is represented as follows:

\[
y = M(d \mid \textbf{w})
\]

In this context, we propose that an authorized auditor has limited access to the trained model $M$. This access enables the auditor to gain insight into the behavior of the model as it processes data $d$. Specifically, the auditor can access the model gradients, denoted as $\nabla_\textbf{w}$, which are calculated as the data sample $d$ passes through the trained model. These gradients provide a detailed view of how the model parameters respond to input data. Gradients can be extracted from each weight $w_i$ in the network, i.e., from different layers at various depths of the model. This ability to access gradients from various layers allows the auditor to analyze the influence of different parts of the model architecture on the final output.

MINT aims to identify whether the given data instance $d$ is part of the training dataset $\mathcal{D}$ used to train model $M$. To achieve this, an authorized entity uses training data $\mathcal{D}$ and external data $\mathcal{E}$ to train a MINT auditing model $T(\cdot | \boldsymbol\uptheta)$. The MINT model is trained with the gradients of the model $M$, $\nabla \textbf{w}$, calculated as training data $\mathcal{D}$ and external data ($\mathcal{E} \notin \mathcal{D}$) are passed through $M$. This ensures that the auditing model captures the audited model's responses to both sets. The MINT model then predicts whether a data sample $d$ belongs to training data $\mathcal{D}$ or external data $\mathcal{E} \notin \mathcal{D}$. MINT models take advantage of the memorization capabilities inherent in machine learning processes. The fundamental elements of MINT are the following.

\noindent \textbf{Audited Model $M$:} a trained Language Model, defined by a set of learned parameters $\textbf{w}$.

\noindent \textbf{Training Dataset $\mathcal{D}$ and External Dataset $\mathcal{E}$:} the dataset used to train model $M$. In this work $\mathcal{D}$ is a collection of text samples used to learn linguistic patterns, word relationships, and contextual meaning related to text classification tasks. The External Data $\mathcal{E}$ is a collection of text samples that is not included in the training dataset ($\mathcal{D}$).     

\noindent \textbf{Gradients $\nabla \textbf{w}$:} a gradient refers to the vector of partial derivatives of a loss function with respect to the model's parameters (weights). Indicates the direction and magnitude of the steepest ascent of the loss function in the parameter space. During training, the gradients are computed via backpropagation and are used in gradient-based optimization algorithms (such as popular Stochastic Gradient Descent, Adam, or RMSprop) to update the model's parameters in the direction that minimizes the loss. Gradients provide crucial information about how each parameter should be adjusted to reduce prediction error and improve learning. The gradients of the model weights $\textbf{w}$ were computed with respect to the input samples $d$. The hypothesis underlying this work is that the gradients of samples $\mathcal{D}$ will exhibit patterns different from those of the gradients obtained from samples $\mathcal{E}$. 

\noindent \textbf{MINT Model $T(\nabla \textbf{w} | \boldsymbol\uptheta)$ (Auditing Tool):} a model trained to perform the membership inference task, using input gradients to predict whether a given sample $d$ belongs to the training dataset $\mathcal{D}$ or to external data $\mathcal{E}$. The architecture of the MINT model is a fully connected neural network with 3 layers, consisting of dense layers with 256, 128, and 64 neurons (ReLU), followed by a sigmoid output layer with one unit. The input size of the MINT model depends on the number of gradients of the audited model. In order to reduce the size of the input space, in our experiments we used the gradients of the first two layers of the audited model. It uses binary crossentropy as the loss function and is optimized with Adam. Training runs for 100 epochs with a batch size of 64.

\section{Experimental Framework} \label{experimentalframework}

We first introduce the text databases used in this study, which include six different datasets representing three different NLP tasks: one for Topic Classification, one for Entity Linking, and four for Sentiment Analysis. We will provide an overview of their content and structure. Following this, we will present the audited Language Models, detailing their characteristics and configurations. Subsequently, we will outline our experimental protocol.

\subsection{Databases} \label{databases}

\noindent \textbf{AGNews.} \cite{agnews} Topic Classification database with 127,600 news articles categorized into: world, sports, business, and science / tech.

\noindent \textbf{DBPedia.} \cite{dbpedia} Entity Linking database derived from \textit{Wikipedia}, used for categorizing articles into predefined topics.  A taxonomy dataset with 342,782 Wikipedia articles classified into 70 hierarchical categories.

\noindent \textbf{IMDB.} \cite{imdb} The IMDB dataset is a text classification database used for sentiment analysis. Includes 50,000 movie reviews, equally divided into positive and negative sentiments.

\noindent \textbf{Steam Reviews.} \cite{lin2019empirical} The Steam Reviews dataset, consisting of 392,200 reviews, is a text classification database used for sentiment analysis of user reviews on the \textit{Steam} platform. Each review is labeled as either positive or negative, based on the sentiment expressed by the user.

\noindent \textbf{Sentiment140.} \cite{go2009twitter} The Sentiment140 dataset is a text classification database used for sentiment analysis, particularly for classifying tweets into positive and negative sentiments. It contains 1.6 million labeled tweets, with each tweet labeled as either positive, negative, or neutral based on the sentiment it conveys.

\noindent \textbf{TwitterSentiment.} \cite{kedia2020hands} The Twitter Sentiment dataset, consisting of 76,400 tweets, is a text classification database used for analyzing sentiment in \textit{Twitter} posts. Each tweet is labeled into one of four categories: positive, negative, neutral, or irrelevant, based on the sentiment conveyed in the text.

\begin{table*}[t!]
\centering
\resizebox{\textwidth}{!}{%
\begin{tabular}{lccccccc}
\hline
\textbf{}                & \multicolumn{7}{c}{\textbf{Accuracy of Audited Language Models for Different Text Classification Benchmarks}}                                                                                       \\ \hline
\textbf{Databases}       & \textbf{BLSTM} & \textbf{BERT} & \textbf{DistilBERT} & \textbf{XLNet} & \textbf{XLNet-Large} & \textbf{ELECTRA} & \textbf{ELECTRA-Large} \\ \hline
AgNews (177 tokens)                   & 91.17\%       & 91.30\%       & 93.79\%              & 88.98\%       & 91.54\%             & 88.09\%          & 92.73\%                 \\
DBPedia (172 tokens)                  & 96.54\%       & 97.45\%       & 97.55\%              & 89.46\%       & 92.45\%             & 90.78\%          & 91.25\%                 \\
IMDB (165 tokens)                    & 86.34\%       & 90.16\%       & 89.30\%              & 88.21\%       & 90.17\%             & 84.24\%          & 87.52\%                 \\
SteamReviews (50 tokens)            & 85.67\%       & 86.11\%       & 86.29\%              & 77.35\%       & 81.46\%             & 81.36\%          & 83.45\%                 \\
Sentiment140 (170 tokens)            & 76.19\%       & 78.23\%       & 79.25\%              & 71.96\%       & 72.45\%             & 74.44\%          & 73.24\%                 \\
TwitterSentiment (165 tokens)        & 86.82\%       & 78.37\%       & 84.99\%              & 74.62\%       & 79.41\%             & 70.23\%          & 72.81\%                 \\
\hline
\end{tabular}
}
\caption{Classification accuracy of the audited language models evaluated on multiple text classification benchmarks (maximum number of tokens per sample in brackets). Classification accuracy (in percentages) achieved by each model: BLSTM, BERT, DistilBERT, XLNet, XLNet-Large, ELECTRA, and ELECTRA-Large.}
\label{tab:acc}
\end{table*}

\subsection{Audited Language Models} \label{languagemodels}

Different models $M$ are evaluated to analyze whether variations in depth and complexity of the audited model affect MINT performance. The Language Models evaluated in this study are:  \textbf{BERT} \cite{devlin2018bert}, \textbf{DistilBERT} \cite{sanh2019distilbert}, \textbf{XLNet} and \textbf{XLNet-Large} \cite{yang2019xlnet}, \textbf{ELECTRA} and \textbf{ELECTRA-Large} \cite{clark2020electra}, and a \textbf{Custom BLSTM}.  

\subsection{Experimental Protocol} \label{experimentalprotocol}

To evaluate the MINT model in various scenarios, the models listed in Section \ref{languagemodels} were trained using the databases outlined in Section \ref{databases}. Each language model utilizes its own tokenizer throughout the process. All language models have been trained for a total of 50 epochs with a batch size of 512, using Adam as the optimizer. Data separation into training and test sets was carried out by assigning 65\% for training and 35\% for testing. The results of the trained models for the different tasks are presented in Table \ref{tab:acc}.

Once the models were trained, we obtained the gradients of a subset of samples from the dataset, including the fragments $\mathcal{D}$ and $\mathcal{E}$. The goal is to make the training subset $\mathcal{D}$ as small as possible to minimize the number of samples that the developer of the audited model must provide. Consequently, the number of samples is gradually reduced throughout the process (as we will see in the next sections).

This approach is applied in two experimental setups: \textbf{Intra-Database Evaluation (1-vs-1)} and \textbf{Mixed-Database Evaluation (1-vs-6)}, as described in Section 5. 
In the Intra-Database Evaluation setup (1-vs-1), gradients are directly extracted from the $\mathcal{D}$ and $\mathcal{E}$ splits of the same dataset used during training, with a focus on comparing the dataset to itself. In this experimental setup, training samples from $\mathcal{D}$ and $\mathcal{E}$ consist of text from a similar domain. 
For the Mixed-Database Evaluation (1-vs-6), gradients $\mathcal{E}$ are extracted not only from the test samples of the original dataset but also from external text sources, simulated using data from other databases with distinct thematic domains. These additional data are incorporated to better replicate real-world scenarios.

The selected samples, which include both $\mathcal{D}$ and $\mathcal{E}$ samples, are classified based on two classes. This classification is then used as input for the MINT model, which predicts whether a sample $d$ belongs to $\mathcal{D}$ or $\mathcal{E}$. By extracting and analyzing gradients, this process identifies patterns in the behavior of the model, providing a robust framework for auditing machine learning models and enhancing transparency. After extraction, the gradients are flattened and those corresponding to the last three layers are selected to be used as input for the MINT model. The MINT model itself is trained for a total of 100 epochs with a batch size of 32 for all models. Data separation is carried out by assigning 65\% for training and 35\% for testing. 

\begin{table}[t!]
\centering
\Huge
\resizebox{\columnwidth}{!}{
\begin{tabular}{lccccc}
\hline
\multicolumn{1}{l}{\multirow{2}{*}{\textbf{Databases}}} & \multicolumn{5}{c}{\textbf{Number of MINT Samples ($\mathcal{D}$ / $\mathcal{E}$)}} \\
\multicolumn{1}{c}{}                                    & \textbf{2.5K/2.5K} & \textbf{2.25K/2.25K} & \textbf{1.5K/1.5K} & \textbf{1.25K/1.25K} & \textbf{750/750} \\\hline

\multicolumn{6}{c}{\textbf{BSLTM (13M parameters)}} \\\hline
AgNews            & 0.9224  & 0.9199  & 0.8702  & 0.8051  & 0.7212  \\
DBPedia           & 0.9257  & 0.9261  & 0.8560  & 0.7795  & 0.7353  \\
IMDB              & 0.9352  & 0.9142  & 0.8942  & 0.8513  & 0.8120  \\
Sentiment140      & 0.9273  & 0.9017  & 0.8252  & 0.7974  & 0.7061  \\
TwitterSentiment  & 0.9482  & 0.9351  & 0.8990  & 0.7983  & 0.7628  \\
SteamReviews      & 0.9355  & 0.9271  & 0.8479  & 0.8113  & 0.7596  \\\hline

\multicolumn{6}{c}{\textbf{BERT (110M parameters)}} \\\hline
AgNews            & 0.9566  & 0.9378  & 0.8841  & 0.8591  & 0.7393  \\
DBPedia           & 1.0000  & 1.0000  & 0.9982  & 0.9979  & 0.9918  \\
IMDB              & 0.9956  & 0.9982  & 0.9906  & 0.9729  & 0.9614  \\
Sentiment140      & 1.0000  & 1.0000  & 0.9999  & 0.9997  & 0.9965  \\
TwitterSentiment  & 0.9996  & 1.0000  & 0.9937  & 0.9996  & 0.9893  \\
SteamReviews      & 0.9921  & 0.9909  & 0.9527  & 0.9489  & 0.8993  \\\hline

\multicolumn{6}{c}{\textbf{DistilBERT (66M parameters)}} \\\hline
AgNews            & 0.9851  & 0.9797  & 0.9600  & 0.9101  & 0.8343  \\
DBPedia           & 0.9994  & 0.9912  & 0.9934  & 0.9882  & 0.9355  \\
IMDB              & 0.9575  & 0.9793  & 0.9372  & 0.8597  & 0.8269  \\
Sentiment140      & 0.9561  & 0.9225  & 0.8912  & 0.8640  & 0.7546  \\
TwitterSentiment  & 0.9968  & 0.9974  & 0.9147  & 0.8641  & 0.8544  \\
SteamReviews      & 0.9824  & 0.9892  & 0.9668  & 0.9404  & 0.8757  \\\hline

\multicolumn{6}{c}{\textbf{ELECTRA (110M parameters)}} \\\hline
AgNews            & 1.0000  & 1.0000  & 0.9960  & 0.9992  & 0.9945  \\
DBPedia           & 0.9983  & 0.9997  & 0.9965  & 0.9999  & 0.9955  \\
IMDB              & 0.9626  & 0.9802  & 0.9525  & 0.9532  & 0.8752  \\
Sentiment140      & 1.0000  & 1.0000  & 0.9995  & 0.9979  & 0.9818  \\
TwitterSentiment  & 1.0000  & 1.0000  & 0.9988  & 0.9988  & 0.9908  \\
SteamReviews      & 0.9993  & 0.9991  & 0.9976  & 0.9875  & 0.9618  \\\hline

\multicolumn{6}{c}{\textbf{ELECTRA-Large (335M parameters)}} \\\hline
AgNews            & 0.9991  & 0.9954  & 1.0000  & 0.9852  & 0.9751  \\
DBPedia           & 0.9932  & 0.9997  & 0.9974  & 1.0000  & 0.9942  \\
IMDB              & 1.0000  & 1.0000  & 0.9912  & 0.9923  & 0.9823  \\
Sentiment140      & 1.0000  & 1.0000  & 1.0000  & 0.9998  & 0.9900  \\
TwitterSentiment  & 0.99975  & 0.9975  & 1.0000  & 0.9995  & 0.9895  \\
SteamReviews      & 1.0000  & 1.0000  & 1.0000  & 0.9944  & 0.9884  \\\hline

\multicolumn{6}{c}{\textbf{XLNet (110M parameters)}} \\\hline
AgNews            & 1.0000  & 0.9975  & 1.0000  & 0.9984  & 0.9541  \\
DBPedia           & 0.9987  & 1.0000  & 1.0000  & 1.0000  & 0.9545  \\
IMDB              & 0.9941  & 0.9998  & 0.9954  & 0.9754  & 0.9432  \\
Sentiment140      & 0.9952  & 0.9985  & 0.9852  & 0.9621  & 0.9760  \\
TwitterSentiment  & 1.0000  & 1.0000  & 0.9963  & 0.9900  & 0.9569  \\
SteamReviews      & 1.0000  & 1.0000  & 0.9952  & 0.9754  & 0.9784  \\\hline

\multicolumn{6}{c}{\textbf{XLNet-Large (340M parameters)}} \\\hline
AgNews            & 0.9993  & 1.0000  & 0.9824  & 0.9952  & 0.9421  \\
DBPedia           & 1.0000  & 0.9999  & 1.0000  & 0.9870  & 0.9447  \\
IMDB              & 0.9977  & 0.9903  & 0.9814  & 0.9710  & 0.9554  \\
Sentiment140      & 0.9982  & 0.9945  & 0.9964  & 0.9927  & 0.9852  \\
TwitterSentiment  & 0.9997  & 0.9984  & 0.9856  & 0.9875  & 0.9510  \\
SteamReviews      & 1.0000  & 1.0000  & 1.0000  & 0.9998  & 0.9935  \\\hline
\end{tabular}}
\caption{MINT AUC scores for different datasets and $\mathcal{D}$ and $\mathcal{E}$ sizes across the Language Models evaluated for the Intra-Database setup (1-vs-1).}
\label{tab:unified_results}
\end{table}

\section{Experiments and Results} \label{experiments}

\subsection{Intra-Database Evaluation} \label{isolated}
In the Intra-Database evaluation configuration (1-vs-1), both $\mathcal{D}$ and $\mathcal{E}$ are from the same dataset to assess the model's performance on data with similar characteristics, such as the text domain and sources. Using the same dataset ensures that the samples come from a similar context and present a common distribution. This approach removes potential latent variables that could arise if data from different sources or with varying features were used, making the comparison more consistent. This setup helps to guarantee that the results are not influenced by external factors or inconsistencies, allowing for a clearer understanding of how well the model can classify and generalize within the given dataset. The results of the Intra-Database evaluation are presented in Table \ref{tab:unified_results}.

The results in Table \ref{tab:unified_results} show that the model effectively distinguishes between a sample $d$ that belongs to $\mathcal{D}$ and a sample $d$ that belongs to $\mathcal{E}$ that has the same context and distribution, highlighting its strong generalization within the dataset. The AUC values in the tables demonstrate robust performance under most conditions. High AUC scores indicate the model's ability to reliably differentiate between classes, reflecting its accuracy with sufficient data. As expected, the best results for all models are obtained when the number of samples to train the MINT Model $T$ is higher. In contrast, as the number of MINT training samples decreases, the AUC gradually decreases, reaching its minimum value when the samples are 750 each (750 samples from $\mathcal{D}$ and 750 samples from $\mathcal{E}$). In most of the experiments, 3,000 training samples (1,500 from each class $\mathcal{D}$ and $\mathcal{E}$), was enough to obtain very competitive performances.

The limited sample size likely prevents an adequate representation of the data distribution, impedes the learning and generalization of the model, and its ability to distinguish between samples belonging to $\mathcal{E}$ and samples belonging to $\mathcal{D}$. Furthermore, we can observe that as the complexity of the model increases, the results improve, with the highest AUC values obtained using more complex architectures with millions of parameters. 

\begin{figure}[t!]

\centering
\includegraphics[width=0.45\textwidth]{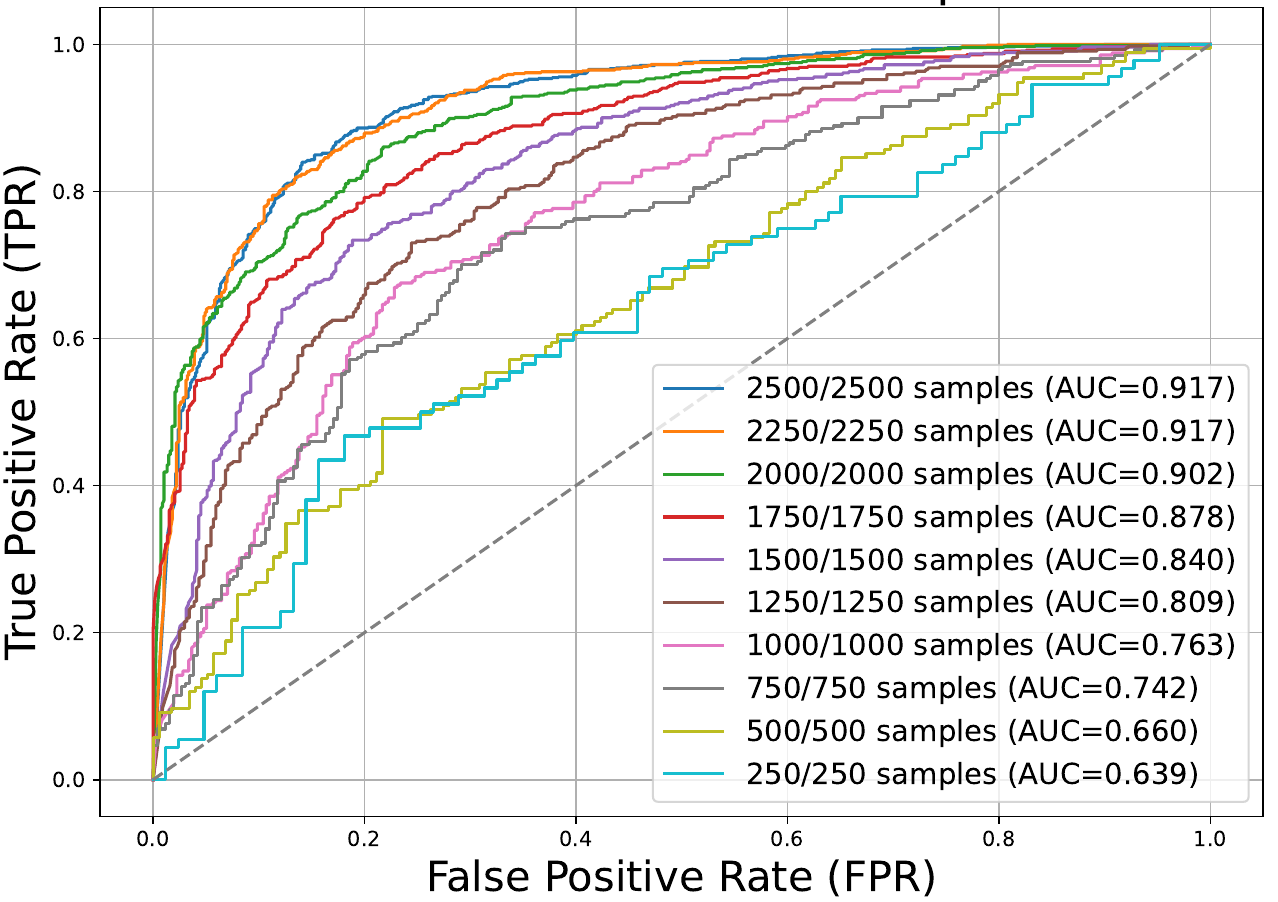}
\caption{ROC curves for the BSLTM model on DBPedia, varying the size of $\mathcal{D}$ and $\mathcal{E}$. (Intra-Database.)}
\label{fig:ROCbsltmdbpedia}
\end{figure}

\begin{figure}[t!]
\centering
\includegraphics[width=0.45\textwidth]{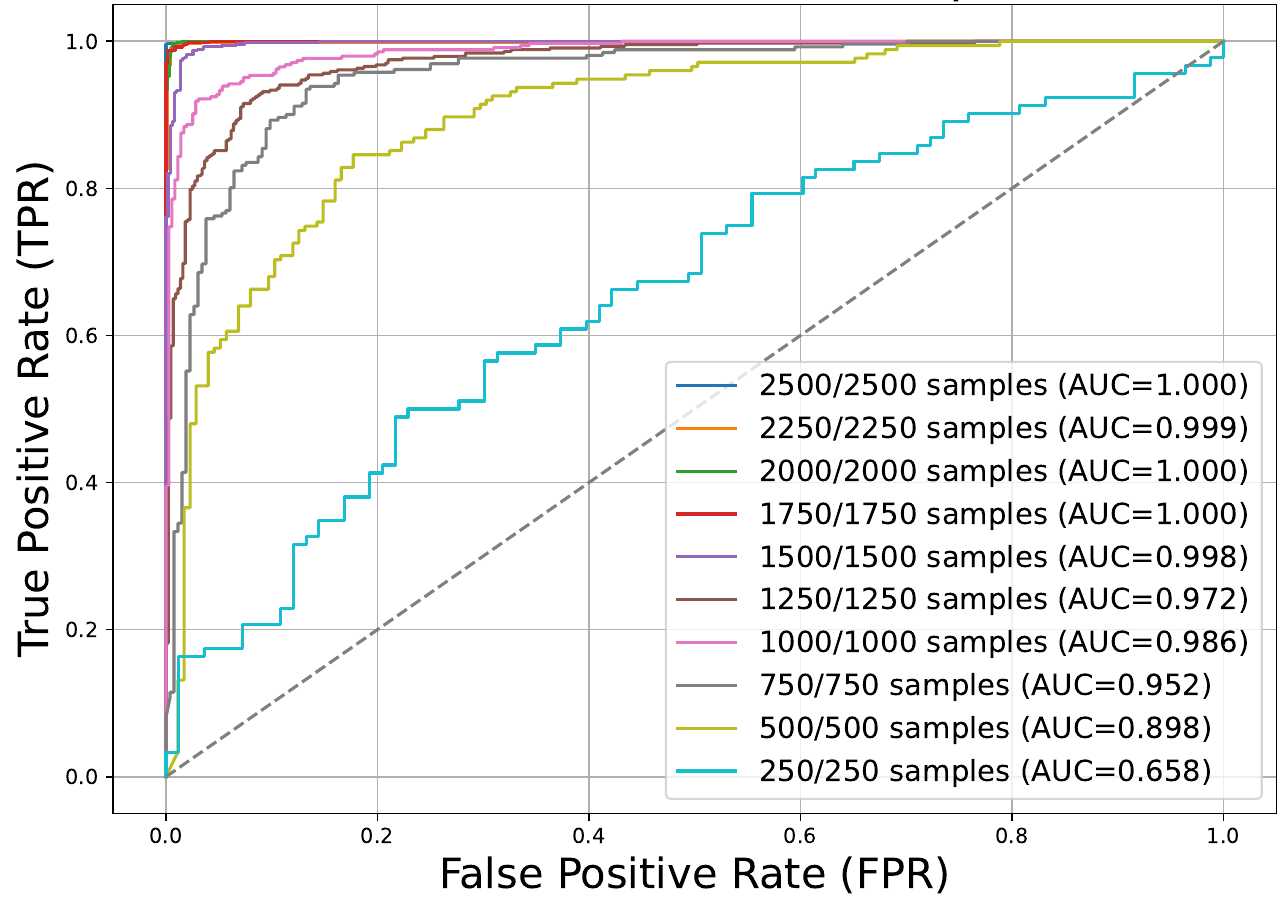}
\caption{ROC curves for the XLN model on DBPedia, varying the size of $\mathcal{D}$ and $\mathcal{E}$. (Intra-Database.)}
\label{fig:ROCXLNbpedia}
\end{figure}

To better visualize the effect of model complexity and sample size, two architectures are chosen: customized BSLTM, the simplest with the fewest parameters, and XLN, one of the most complex. Furthermore, the number of MINT training samples is progressively increased. Figures \ref{fig:ROCbsltmdbpedia} and \ref{fig:ROCXLNbpedia} present the ROC curves for these two architectures and a different number of MINT training samples ($\mathcal{D}$ and $\mathcal{E}$). The results show the trade-off between the number of samples and the performances.

\subsection{Mixed-Database Evaluation} \label{composite}

In the Mixed-Database evaluation setup (1-vs-6), the training data $\mathcal{D}$ were taken from one dataset, while the external data $\mathcal{E}$ come from all six datasets, including unseen samples from $\mathcal{D}$ (i.e., samples from $\mathcal{D}$ not used to train the MINT model $T$). By introducing more datasets, the evaluation reflects real-world scenarios and how the model adapts to different data distributions and contexts, providing insight into its ability to generalize beyond the specific nature of $\mathcal{D}$, as shown in Table \ref{tab:unified_results1}.

\begin{table}[t!]
\centering
\Huge
\resizebox{\columnwidth}{!}{
\begin{tabular}{lccccc}
\hline
\multicolumn{1}{l}{\multirow{2}{*}{\textbf{Databases}}} & \multicolumn{5}{c}{\textbf{Number of MINT Samples ($\mathcal{D}$ / $\mathcal{E}$)}} \\
\multicolumn{1}{c}{}                                    & \textbf{2.5K/2.5K} & \textbf{2.25K/2.25K} & \textbf{1.5K/1.5K} & \textbf{1.25K/1.25K} & \textbf{750/750} \\ \hline

\multicolumn{6}{c}{\textbf{BSLTM (13M parameters)}} \\ \hline
AgNews                                                  & 0.9335             & 0.9056               & 0.8852             & 0.8521               & 0.8107           \\ 
DBPedia                                                 & 0.9242             & 0.9123               & 0.8924             & 0.8856               & 0.8700           \\ 
IMDB                                                    & 0.9245             & 0.9256               & 0.8879             & 0.8421               & 0.7996           \\ 
Sentiment140                                            & 0.9215             & 0.9352               & 0.8702             & 0.8615               & 0.8120           \\ 
TwitterSentiment                                        & 0.9321             & 0.9351               & 0.9120             & 0.8623               & 0.8550           \\ 
SteamReviews                                           & 0.9451             & 0.9387               & 0.8975             & 0.8569               & 0.7950           \\ \hline

\multicolumn{6}{c}{\textbf{BERT (110M parameters)}} \\ \hline
AgNews                                                  & 0.9697             & 0.9716               & 0.9697             & 0.9510               & 0.9333           \\
DBPedia                                                 & 0.9996             & 1.0000                    & 1.0000                  & 0.9999               & 0.9997           \\
IMDB                                                    & 0.9966             & 0.9945               & 0.9797             & 0.9775               & 0.9560           \\
Sentiment140                                            & 0.9994             & 1.0000                    & 0.9994             & 0.9968               & 0.9910           \\
TwitterSentiment                                        & 0.9993             & 0.9987               & 0.9986             & 0.9930               & 0.9734           \\
SteamReviews                                           & 0.9880             & 0.9858               & 0.9598             & 0.9519               & 0.9097           \\ \hline

\multicolumn{6}{c}{\textbf{DistilBERT (66M parameters)}} \\ \hline
AgNews                                                  & 1.0000                  & 0.9999               & 0.9990             & 0.9998               & 0.9919           \\
DBPedia                                                 & 0.9989             & 0.9998               & 0.9989             & 0.9990               & 0.9996           \\
IMDB                                                    & 0.9779             & 0.9320               & 0.9020             & 0.9268               & 0.8639           \\
Sentiment140                                            & 0.9381             & 0.9203               & 0.8897             & 0.8241               & 0.8196           \\
TwitterSentiment                                        & 0.9982             & 0.9988               & 0.9943             & 0.9938               & 0.9858           \\
SteamReviews                                           & 0.9498             & 0.9831               & 0.9350             & 0.9481               & 0.9050           \\ \hline

\multicolumn{6}{c}{\textbf{ELECTRA (110M parameters)}} \\ \hline
AgNews                                                  & 0.9987             & 0.9972               & 0.9871             & 0.9821               & 0.9721           \\
DBPedia                                                 & 0.9989             & 0.9998               & 0.9989             & 0.9996               & 0.9998           \\
IMDB                                                    & 0.9970             & 0.9960               & 0.9943             & 0.9940               & 0.9786           \\
Sentiment140                                            & 0.9997             & 0.9992               & 0.9976             & 0.9978               & 0.9844           \\
TwitterSentiment                                        & 0.9998             & 0.9998               & 0.9994             & 0.9985               & 0.9974           \\
SteamReviews                                           & 0.9983             & 0.9978               & 0.9886             & 0.9847               & 0.9647           \\ \hline

\multicolumn{6}{c}{\textbf{ELECTRA-Large (335M parameters)}} \\ \hline
AgNews                                                  & 1.0000                  & 1.0000                    & 0.9871             & 0.9821               & 0.9721           \\
DBPedia                                                 & 1.0000                  & 1.0000                    & 1.0000                  & 1.0000                    & 0.9998           \\
IMDB                                                    & 0.9970             & 0.9960               & 0.9943             & 0.9940               & 0.9786           \\
Sentiment140                                            & 1.0000                  & 1.0000                    & 0.9906             & 0.9964               & 0.9901           \\
TwitterSentiment                                        & 1.0000                  & 1.0000                    & 1.0000                  & 0.9992               & 0.9984           \\
SteamReviews                                           & 1.0000                  & 0.9995               & 0.9977             & 0.9901               & 0.9896           \\ \hline

\multicolumn{6}{c}{\textbf{XLNet (110M parameters)}} \\ \hline
AgNews                                                  & 0.9956             & 0.9954               & 0.9940             & 0.9967               & 0.9359           \\
DBPedia                                                 & 0.9997             & 0.9993               & 0.9988             & 0.9986               & 0.9854           \\
IMDB                                                    & 0.9814             & 0.9966               & 0.9821             & 0.9662               & 0.9418           \\
Sentiment140                                            & 0.9979             & 0.9929               & 0.9778             & 0.9740               & 0.9715           \\
TwitterSentiment                                        & 1.0000                  & 0.9912                    & 1.0000                  & 0.9972               & 0.9994           \\
SteamReviews                                           & 0.9998             & 0.9999               & 0.9991             & 0.9980               & 0.9820           \\ \hline

\multicolumn{6}{c}{\textbf{XLNet-Large (340M parameters)}} \\ \hline
AgNews                                                  & 1.0000             & 1.0000               & 1.0000             & 0.9966               & 0.9456           \\
DBPedia                                                 & 1.0000             & 0.9993               & 1.0000             & 0.9986               & 0.9854           \\
IMDB                                                    & 0.9975             & 1.0000              & 0.9954             & 0.974               & 0.961          \\
Sentiment140                                            & 0.9984             & 0.9973              & 0.9954             & 0.9841               & 0.9815           \\
TwitterSentiment                                        & 1.0000                  & 0.9915                    & 1.0000                  & 0.9947              & 0.9965           \\
SteamReviews                                           & 1.0000             & 1.0000               & 1.0000            & 0.9972              & 0.9920           \\ \hline
\end{tabular}}
\caption{MINT AUC scores for different datasets and $\mathcal{D}$  and $\mathcal{E}$ sizes across the Language Models evaluated for the Mixed-Database setup (1-vs-6).}
\label{tab:unified_results1}
\end{table}

\begin{figure}[t!]
\centering
\includegraphics[width=0.45\textwidth]{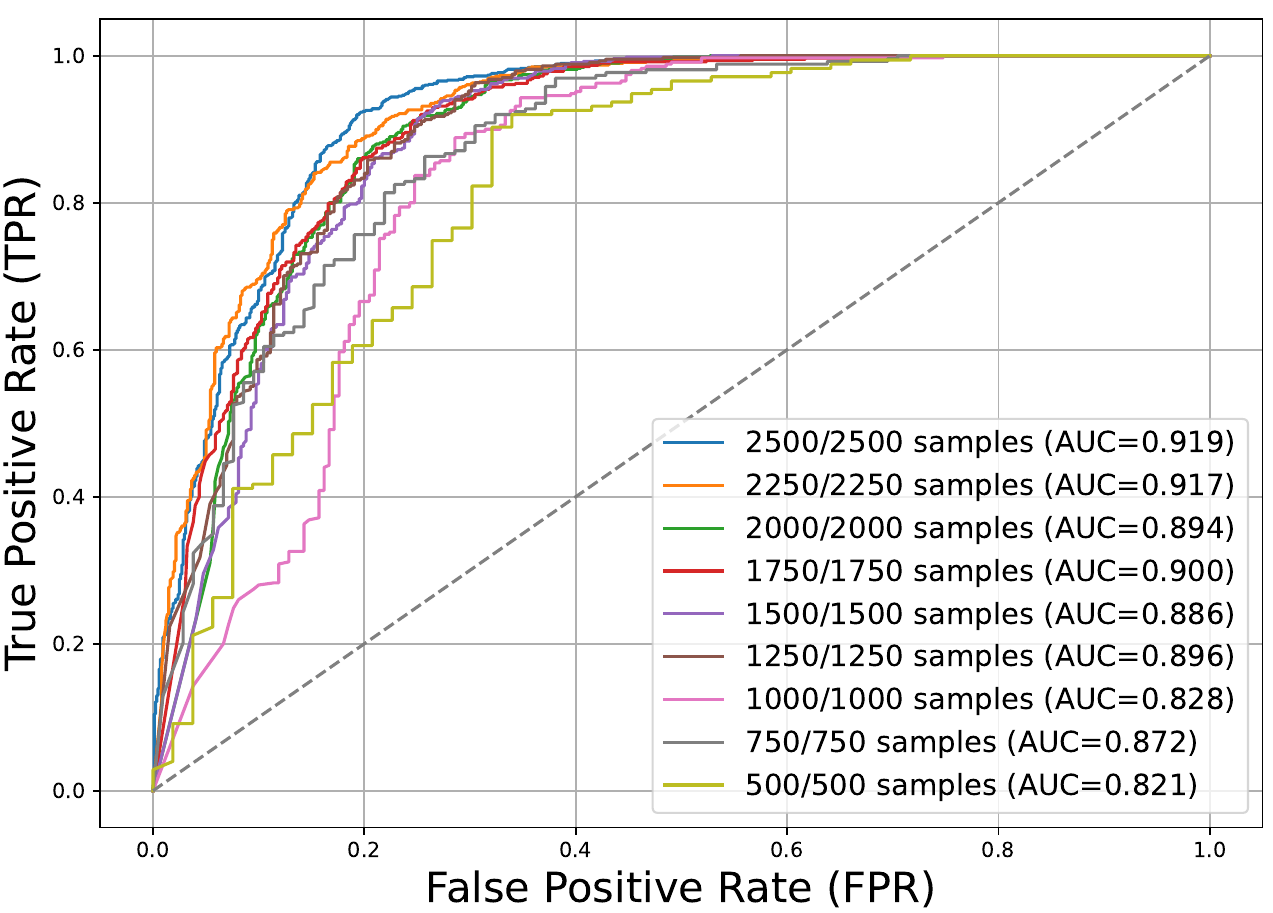}
\caption{ROC curves for the BLSTM model on DBPedia, varying the size of $\mathcal{D}$ and $\mathcal{E}$. (Mixed-Database.)}
\label{fig:blstm1vs6}
\end{figure}

\begin{figure}[t!]
\centering
\includegraphics[width=0.45\textwidth]{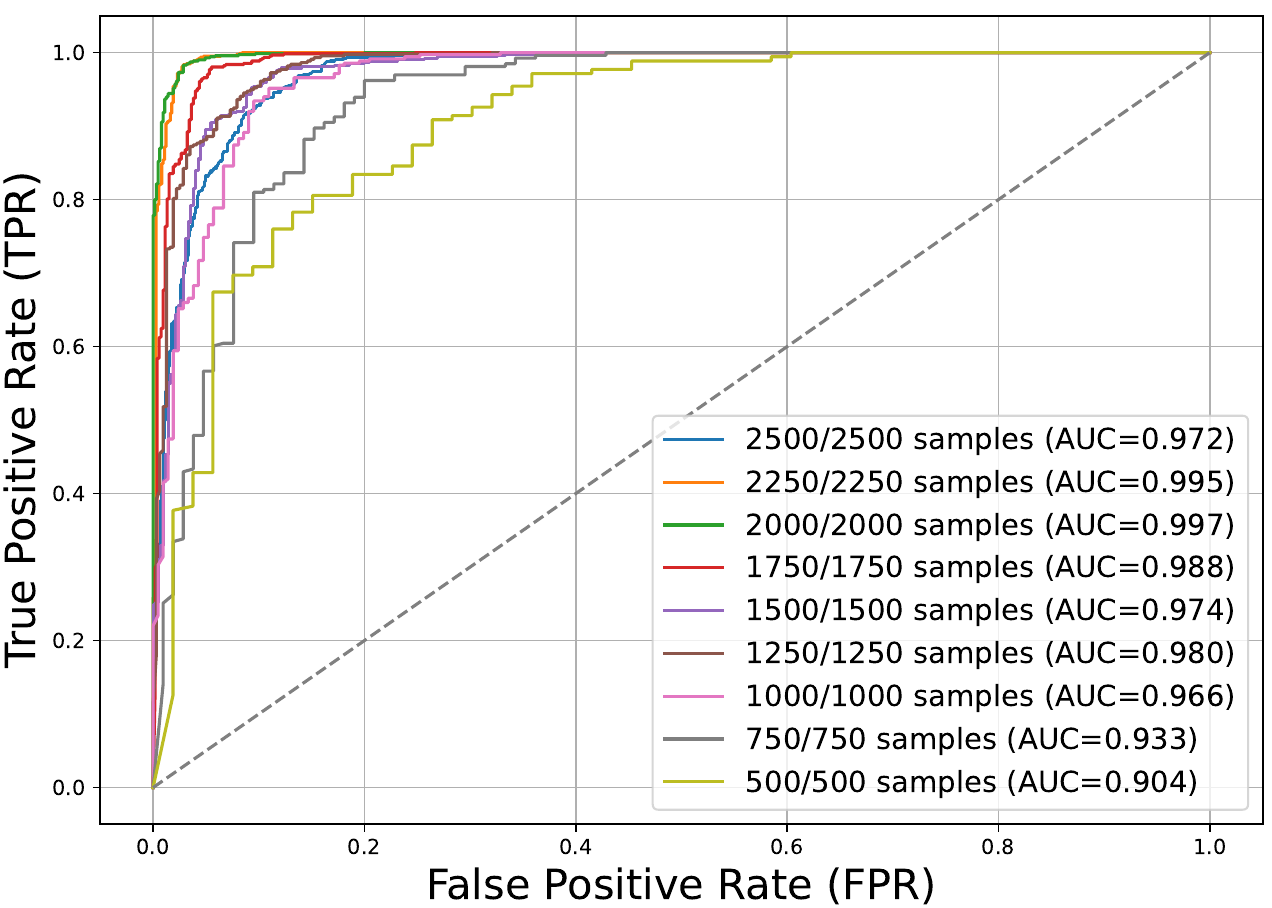}
\caption{ROC curves for the XLN model on DBPedia, varying the size of $\mathcal{D}$ and $\mathcal{E}$. (Mixed-Database.)}
\label{fig:xln1vs6}
\end{figure}

The results follow the same trend as those obtained in Section \ref{isolated}. It can be seen from the high AUC values that the MINT model is able to differentiate between $\mathcal{D}$ and $\mathcal{E}$ from the same database, as well as from different sources. As in the previous section, the effect of model complexity and sample size is analyzed in more detail evaluating two architectures: the customized BSLTM, the simplest with the fewest parameters, and XLN, the most complex model evaluated. The results are presented in Figures \ref{fig:blstm1vs6} and \ref{fig:xln1vs6}. In general, the performance obtained in the Mixed-Database evaluation is higher than the performance obtained in the Intra-Database setup. The different context of the external data helps to better differentiate between the samples $\mathcal{D}$ and $\mathcal{E}$. 

\begin{figure}[t!]
\centering
\includegraphics[width=0.49\textwidth]{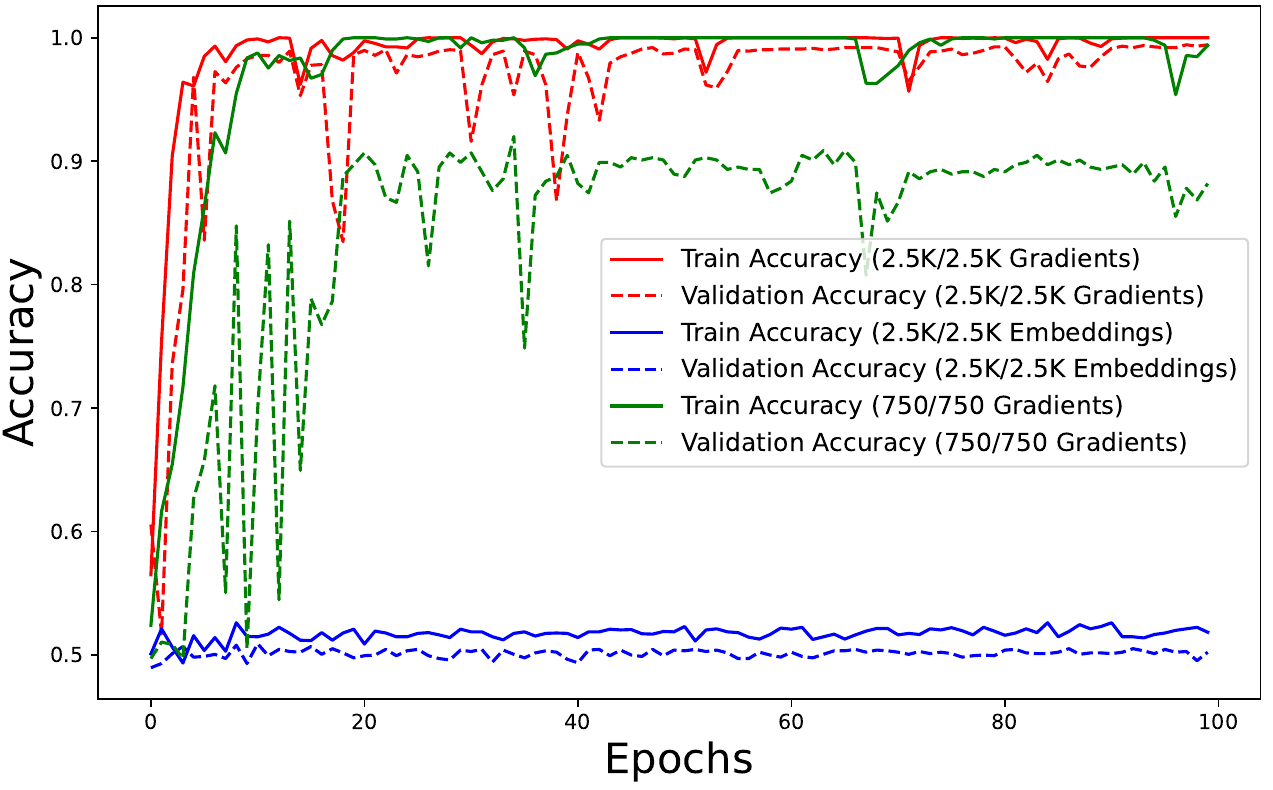}
\caption{Evolution of MINT accuracy over epochs, comparing the previous embedding method \cite{dealcala2024comprehensive}  with our new gradient method, trained using the XLNet model on the Sentiment140 dataset.}
\label{fig:gradvsembeddings}
\end{figure}

Finally, in Table \ref{tab:mintoriginal} we present a comparison with the MINT method previously proposed by \citet{dealcala2024comprehensive}. It is important to note that \citet{dealcala2024comprehensive} proposed a method for Convolutional Neural Networks (CNN) and image input. That method trained the MINT model using CNN embeddings. Our results, shown in Table \ref{tab:mintoriginal} and Fig. \ref{fig:gradvsembeddings} suggest that the MINT model based on CNN embeddings and not network weight gradients like here is not able to separate between $\mathcal{D}$ and $\mathcal{E}$ properly. Note that the Language Models here output a sequence of tokens or a probability distribution over the vocabulary, rather than learned embeddings in the representation space of CNNs.  

All experiments were carried out on a PC with Intel Core i9-14900 5.8 GHz, 64GB RAM, and $2\times$ Nvidia RTX4090 (24GB VRAM). The proposed MINT models can be trained with very limited computational resources, and they can be considered medium-light models.

\begin{table}[t!]
\centering
\small
\begin{tabular}{lcc}
\hline
\textbf{Databases}       & Embeddings* & $\nabla \textbf{w} $ (\textbf{Ours})  \\ \hline
IMDB                     & 0.5152       & 0.9772                    \\
SteamReviews             & 0.5022       & 0.9876                   \\
TwitterSentiment         & 0.4951       & 0.9737                    \\
AgNews                   & 0.5012       & 0.9794                     \\
DBPedia                  & 0.4913       & 0.9907                      \\
Sentiment140             & 0.4871       & 0.9848                   \\
\hline
\end{tabular}
\caption{Average MINT AUC performance of the different models across the six distinct datasets for 2.5K/2.5K samples ($\mathcal{D}$ / $\mathcal{E}$). *We have adapted the MINT method proposed by \citet{dealcala2024comprehensive} for images to our text classification framework.}
\label{tab:mintoriginal}
\end{table}

\section{Conclusions} \label{conclusions}

In this work, a gradient-based Membership Inference Test (gMINT) method was adapted and evaluated to identify data samples used during language model training. The framework demonstrated its effectiveness on six datasets and seven Transformer-based language models. The experimental results revealed that the proposed method achieves high accuracy, with AUC values ranging from 70\% to 99\%, depending on the volume of data and the complexity of the model. 

This study highlights the adaptability of the general MINT approach for detecting if given data were used for training machine learning models in various text classification tasks, showing competitive precision both in intra- and inter-database evaluation. The results reveal that our gradient-based learning methodology has the potential to protect sensitive data and promote ethical AI development. These insights underscore the importance of integrating membership inference frameworks like MINT into machine learning systems to ensure both accountability and trustworthiness.

\section{Limitations and Future Work} \label{limitationss}

To the best of our knowledge, this is the first time MINT has been studied in the context of language models, with certain limitations:

\begin{itemize}
    \item Our experiments focus on text classification tasks. Future research should explore the application of MINT to language models trained for other tasks, particularly increasingly popular generative models. Generative models are characterized by an extremely high number of parameters (in the billions) and a training process that involves vast amounts of data. Our experiments initially suggest that larger number of parameters help to improve the detection of training data.

    \item Our experiments assume a collaborative or, at the very least, non-intrusive stance from the owner of the audited model. Our method requires access to the model's gradients, which are not available in proprietary language models. Future work should investigate vulnerabilities and develop countermeasures for scenarios in which the model owner actively seeks to obfuscate the training data (e.g., introducing small variations in the training data).

    \item Due to space limitations, we have not included experiments that analyze how the number of tokens affects the performance of MINT models. The experiments presented in Sect. \ref{experiments} cover a range of token lengths, varying from $50$ (SteamReview) to $177$ (AgNews). Future research should further investigate the impact of token count on performance, especially in cases where the number of tokens is low.
\end{itemize}

\section{Acknowledgment} 

This study has been supported by the projects BBforTAI (PID2021-127641OB-I00 MICINN/FEDER) and Cátedra ENIA UAM-VERIDAS en IA Responsable (NextGenerationEU PRTR TSI-100927-2023-2). The work G. Mancera is supported by FPI-PRE2022-104499 MICINN/FEDER. The work of D. deAlcala is supported by a FPU Fellowship (FPU21/05785) from the Spanish MIU. The work has been conducted within the ELLIS Unit Madrid. 



\bibliography{acl_latex}
\appendix


\end{document}